\documentclass[sigconf]{acmart}

\renewcommand\footnotetextcopyrightpermission[1]{} 

\usepackage{hyperref}
\usepackage{hyperxmp}
\usepackage{graphicx}
\usepackage{textcomp}
\usepackage{xcolor}
\usepackage{multirow}
\usepackage{ltxtable}
\usepackage{soul}
\usepackage{tikz}
\usepackage{algorithm}
\usepackage{algpseudocode}
\usepackage{colortbl}
\usepackage{bm}
\usepackage{float}
\usepackage{longtable}
\usepackage[caption=false]{subfig}
\usepackage{cellspace}

\newcommand{\cmmnt}[1]{}  

\def\BibTeX{{\rm B\kern-.05em{\sc i\kern-.025em b}\kern-.08em
    T\kern-.1667em\lower.7ex\hbox{E}\kern-.125emX}}

\newcommand\encircle[1]{%
\tikz[baseline=(X.base)] 
  \node (X) [draw, scale=0.75, shape=circle, inner sep=0, fill=black, text=white, minimum size=0em] {\strut #1};}

\begin{document}

\title{HiRISE: High-Resolution Image Scaling for Edge ML via In-Sensor Compression and Selective ROI}

\author{Brendan Reidy$^{\dagger}$, Sepehr Tabrizchi$^\ddagger$, Mohamadreza Mohammadi$^{\dagger}$, Shaahin Angizi$^\S$, Arman Roohi$^\ddagger$, and Ramtin Zand$^{\dagger}$}
\affiliation{
\institution{\small $^\dagger$ Department of Computer Science and Engineering, University of South Carolina, Columbia, SC, USA}
\institution{$^\ddagger$ School of Computing, University of Nebraska–Lincoln, Lincoln, NE, USA}
\institution{$^\S$ Department of Electrical and Computer Engineering, New Jersey Institute of Technology, Newark, NJ, USA}
\city{} \country{}} 
\email{bcreidy@email.sc.edu, stabrizchi2@huskers.unl.edu, mohammm@email.sc.edu,  shaahin.angizi@njit.edu, aroohi@unl.edu, ramtin@cse.sc.edu}

\pagestyle{plain}

\begin{abstract}
With the rise of tiny IoT devices powered by machine learning (ML), many researchers have directed their focus toward compressing models to fit on tiny edge devices. Recent works have achieved remarkable success in compressing ML models for object detection and image classification on microcontrollers with small memory, e.g., 512kB SRAM. However, there remain many challenges prohibiting the deployment of ML systems that require high-resolution images. Due to fundamental limits in memory capacity for tiny IoT devices, it may be physically impossible to store large images without external hardware. To this end, we propose a \ul{hi}gh-\ul{r}esolution
\ul{i}mage \ul{s}caling system for \ul{e}dge ML, called HiRISE, which is equipped with selective region-of-interest (ROI) capability leveraging analog in-sensor image scaling. Our methodology not only significantly reduces the peak memory requirements, but also achieves up to 17.7$\times$ reduction in data transfer and energy consumption.

\end{abstract}

\maketitle

\section{Introduction}
\begin{figure}[] 
\centering
\includegraphics [width=0.95\linewidth]{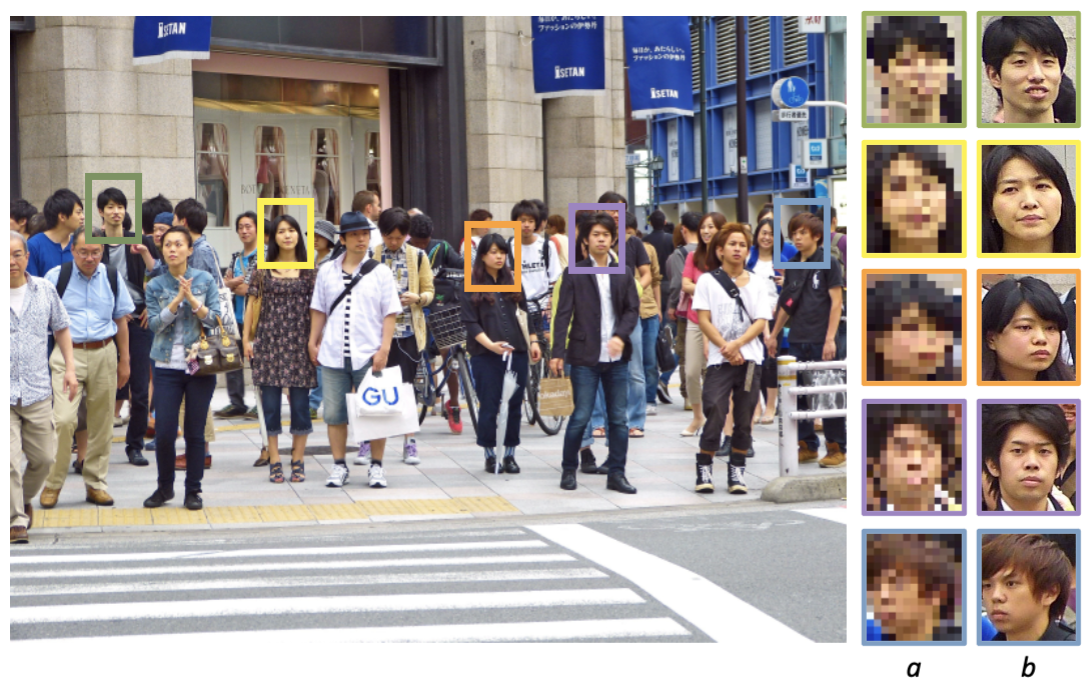}
\vspace{-5mm}
\caption{Example objects of interest with (a) ROI of a 320$\times$240 image compressed in processor, (b) ROI of a 2560$\times$1920 image compressed in sensor to 320$\times$240 using HiRISE system.}
\label{fig:detection_example}
\vspace{-3mm}
\end{figure}

In recent years, machine learning (ML) has witnessed two opposing trends. On one hand, as models have become more capable, they have also grown larger, requiring more resources like memory and energy. On the other hand, we have seen the rise of tiny IoT devices operating with constrained resources, forcing ML models to be smaller. For this reason, many researchers have focused on the daunting task of model compression. Techniques like quantization, pruning, and neural architecture search along with advancements like MobileNets [12] have made it possible to deploy ML models for various applications at the mobile scale without significant degradation in performance. Later works like MCUNet [8] have taken this a step further, making it possible to deploy models at the microprocessor scale (512kB SRAM) without significant performance loss. These advancements have enabled a multitude of new applications and expanded the scope of what is possible at the edge.
Even with these advancements, however, there are still many applications that remain out of reach. Many ML models expect pre-cropped images as input, such as facial recognition models which expect just faces as input. In a live camera feed, however, it is impractical to expect only objects of interest to be in view. For this reason, two-stage approaches are advantageous to be used in which an object detection model first extracts objects of interest, and then the second model performs its task. This allows systems to perform their respective tasks in dynamic, uncontrolled environments. In tiny IoT devices, however, the input image is very small, making the cropped object of interest even smaller. As shown in Fig. \ref{fig:detection_example}(a), for a 320$\times$240 image, we are left with very little information for the second stage model to perform its task. While this resolution might be enough for certain tasks, applications like face recognition rely on rich features such as hair texture, eyes, nose, ears, and more to identify targets [5]. One solution is to simply increase the resolution of the original image, however, for tiny devices where memory is the most scarce resource, the memory required to store the image quickly overtakes that of the model as the new memory bottleneck. 


\begin{figure}[]
\subfloat[Conventional Object Detection System.]{%
  \includegraphics[clip,width=\columnwidth]{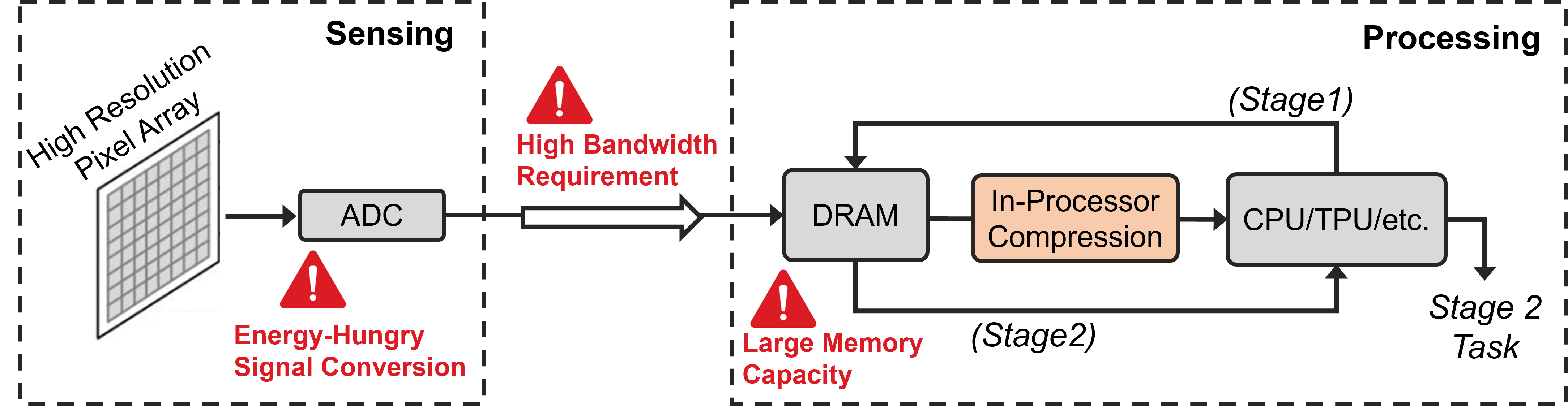}%
}

\subfloat[Proposed Object Detection System.]{%
  \includegraphics[clip,width=\columnwidth]{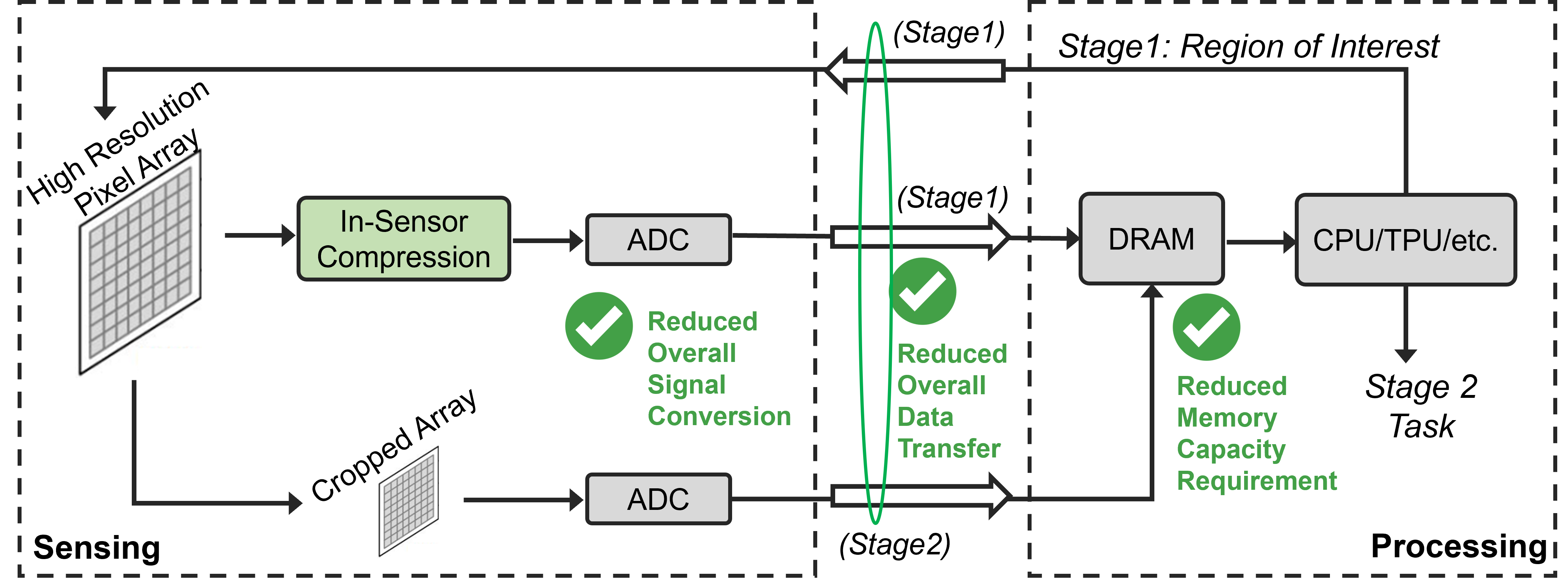}%
}
\vspace{-3mm}
\caption{Project Overview. Our proposed system aims to address three main challenges that make the use of very high-resolution cameras at conventional edge computing systems unfeasible. These challenges include energy-hungry analog-to-digital conversion, substantial data transfer overheads, and intensive memory demands. Our solution includes the implementation of a two-stage object detection model along with an analog in-sensor compression unit.}
\vspace{-4mm}
\label{fig:ProjectOverview}
\end{figure}

As shown in Fig. \ref{fig:ProjectOverview}, the proposed project aims to address three challenges of using very high-resolution cameras at the edge devices; (1) converting 10s of millions of analog pixels to digital signals using energy-hungry analog-to-digital converters (ADCs) demands an energy that might not available in most of the energy-constrained edge devices; (2) transferring 10s of Megabytes of information from the sensing unit to processing unit imposes high bandwidth requirements and significant energy and latency overheads; (3) Storing 10s of Megabytes of information requires a large memory capacity which is not available in most of the resource-constrained edge devices. In this paper, we propose \ul{hi}gh-\ul{r}esolution
\ul{i}mage \ul{s}caling system for \ul{e}dge ML, called HiRISE, to address the aforementioned challenges. To enable the two-stage processing of high-resolution images in HiRISE, we introduce two in-sensor circuits that reduce the peak memory required by the system. The first circuit performs in-sensor average pooling to reduce the size of the image before sending it to the digital hardware. The second circuit extracts high-resolution versions of the ROIs generated from our stage-1 model. We showcase an example of HiRISE in Fig. \ref{fig:detection_example} (b).


Several related works have proposed moving some of the computation to the sensor to reduce memory transfer and computation of ML models [15, 16]. MACSEN [15] processes the first convolutional layer of a Binary convolutional neural network (CNN) in a vision sensor with the correlated double sampling procedure achieving 1000 FPS speed in computation mode. In [14], a processing-in-pixel architecture is designed to support 8-bit activation and weight intended for the first-layer CNN acceleration through pulse modulation. PISA [1] enables convolutional operations in the first layer of Binary CNN by leveraging non-volatile memory to store network weights. In [6], a processing-in-sensor architecture is designed that leverages pixel current and charge-sharing events to enable feature extraction through current-domain MAC operations. In [2], a CNN-based face recognition system is proposed where part of the facial recognition system is moved to the sensor. However, our work is different from these in several ways. First, HiRISE is designed for very high-resolution images. Second, our system only moves the scaling and ROI in the sensor, which makes it easier to generalize HiRISE for other applications.


\section{End-to-End System Architecture}
\begin{figure}[] 
\centering
\includegraphics [width=0.98\linewidth]{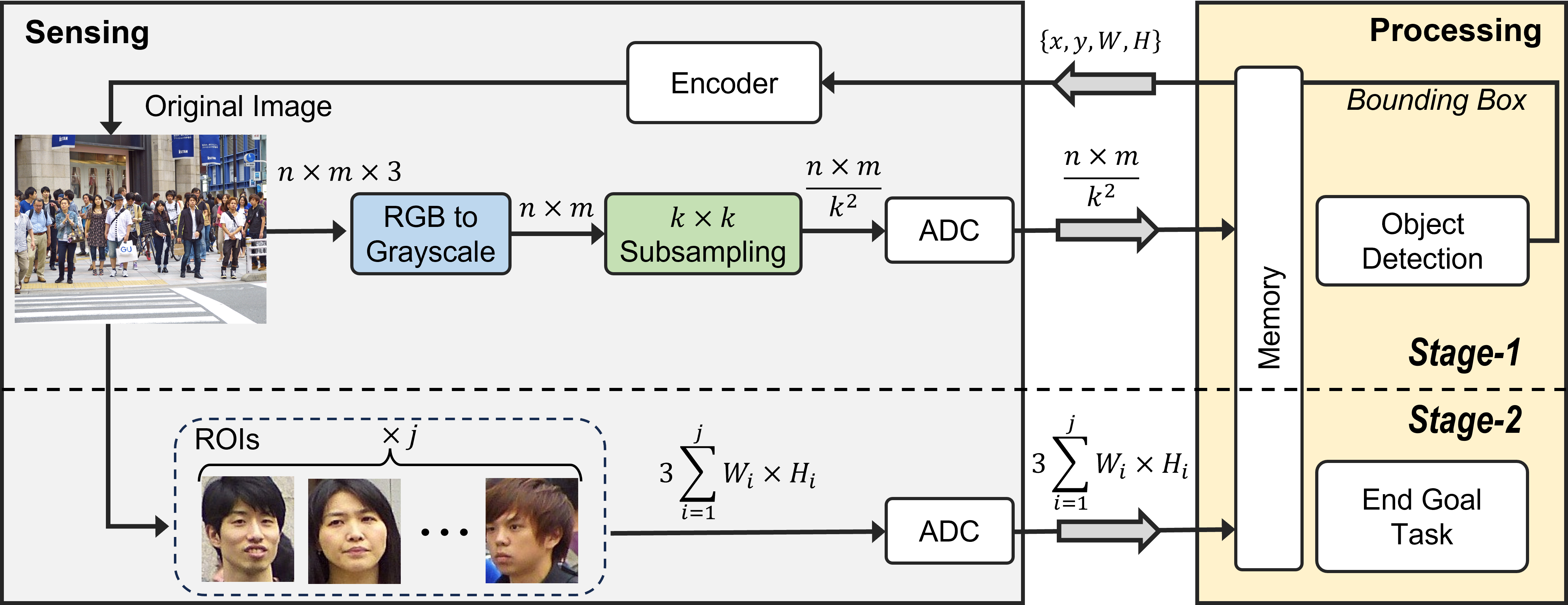}
\vspace{-3mm}
\caption{The overall architecture of the HiRISE end-to-end system including various building blocks and the data transfers between different blocks.}
\label{fig:end-to-end}
\end{figure}

Figure \ref{fig:end-to-end} shows the overall architecture of the proposed HiRISE system that enables processing very high-resolution images for edge devices at a scale not attainable previously. The figure highlights the two stages involved in our proposed methodology to reduce the overall signal conversion and data transfer between the sensing and processing units.

The objective of the first stage is to identify the region of interest (ROI) in the original high-resolution image to avoid unnecessary conversion and transfer of pixels that do not contain valuable information from the sensor to the processor. While having high-quality images is crucial for the final task, e.g. facial recognition from far away, identifying the ROI, e.g. finding a human in the image, does not require very high resolution. This has allowed us to leverage two mechanisms to compress the original images to lower-resolution images and transfer them to the processing unit to identify the ROI. These methods include converting RGB images to grayscale that can realize a $3\times$ compression, followed by a $k\times k$ pooling that reduces the data by a factor of $k^2$, as depicted in Fig. \ref{fig:end-to-end}.

Given that one of the primary aims of this paper is to minimize analog-to-digital signal conversion operations, both compression methods mentioned earlier must occur in the analog domain without undergoing any conversion. This constraint, which is at times overlooked in prior works utilizing in-sensor compression [17], is addressed in this paper through the development of specialized circuitry designed for handling these compression operations in the analog domain, as detailed in the following section. Following the compression process, once the compressed image is transferred to the processing unit, a stage-1 object detection model, trained to identify the ROI in the image, returns the ROI's location ($x, y$) and dimensions ($W, H$) to the sensor. In the second stage, an encoder is employed to select the ROI from the original pixel array based on the bounding box information obtained in the first stage. It then converts the analog pixels into digital signals before transferring them to the processing unit to accomplish the end-goal task, such as object detection and classification.


Table \ref{tab:analytical_model} presents the analytical relationships governing overall data transfer, signal conversion, and memory demands for both conventional object detection systems and the proposed HiRISE system. In the table, $n\times m$ denotes the dimensions of the original image, $P_{ADC}$ represents the precision of the ADC, $k\times k$ corresponds to the subsampling/pooling size, $W \times H$ is the dimensions of the bounding box, and $j$ represents the number of bounding boxes. The goal of the HiRISE system is to fulfill the following three conditions:

\begin{equation}
    D_{new} = D1_{S\rightarrow P} + D1_{P\rightarrow S} + D2_{S\rightarrow P} \ll D_{old}
\end{equation}

\begin{equation}
    Mem_{new} = max(M1_{S\rightarrow P}, M2_{S\rightarrow P}) \ll Mem_{old}
\end{equation}

\begin{equation}
   C_{new} = C1_{S\rightarrow P} + C2_{S\rightarrow P} \ll C_{old}
\end{equation}

\begin{table*}[]
\caption{Analytical relations for data transfer, memory capacity, and signal conversion for HiRISE vs conventional method.}
\vspace{-3mm}
\begin{tabular}{llccc}
\hline
\multicolumn{2}{l}{}                                 & Data Transfer                                                               & Memory Capacity                                                             & ADC                                                  \\ \hline
\multicolumn{2}{l}{Conventional}                     & $D_{old}=(n\times m \times 3).P_{ADC}$     & $Mem_{old}=(n\times m \times 3).P_{ADC}$     & $C_{old}=(n\times m \times 3)$      \\ \hline
\multirow{3}{*}{HiRISE} & \multirow{2}{*}{Stage-1} & $D1_{S\rightarrow P}=(n\times m)/k^2.P_{ADC}$                        & $M1_{S\rightarrow P}=(n\times m)/k^2.P_{ADC}$               &$C1_{S\rightarrow P}=(n\times m)/k^2$ \\
                          &                          & $D1_{P\rightarrow S}=j.(4 \times Words)$                                    & $M1_{P\rightarrow S}=j.(4 \times Words)$                                   & 0                                                    \\ \cline{2-5} 
                          & Stage-2                  & $D2_{S\rightarrow P} =3P_{ADC}. (\sum^{j}_{i=1} (W_{i}\times H_i))$ & $M2_{S\rightarrow P} =3P_{ADC}. (\sum^{j}_{i=1} (W_{i}\times H_i))$  & $C2_{S\rightarrow P} =3\sum^{j}_{i=1} (W_{i}\times H_i)$  \\ \hline
\end{tabular}
\label{tab:analytical_model}
\end{table*}

In equation (1), the $D1_{P\rightarrow S}$ is the total data transfer required for transferring the dimension and coordinates of ROI boxes from processor to sensor that is typically significantly smaller than the $D1_{S\rightarrow P}$, which represents the data transfer required for sending the compressed images from sensor to processor. In addition, $D2_{S\rightarrow P}$ is the intersection over the union of all the ROI boxes transferred from the sensor to the processor for the end-goal task. In equation (2), $M1_{S\rightarrow P}$ is the total memory required for storing the compressed image in stage 1, and $M2_{S\rightarrow P}$ is the total memory required for storing the ROI boxes in stage 2. The memory capacity required in the processing unit should be large enough to fit the maximum of $M1_{S\rightarrow P}$ and $M2_{S\rightarrow P}$, as the compressed image data used for the stage 1 object detection does not need to remain in the memory for stage 2. Finally in equation (3), $C1_{S\rightarrow P}$ and $C2_{S\rightarrow P}$ are the total amount of ADC conversions required for converting the compressed image in the first stage and the  ROIs in the second stage, respectively.



In section 4, we comprehensively investigate the capability of the proposed HiRISE system to achieve the aforementioned conditions using various real-world scenarios and datasets.

\section{In-Sensor Compression Circuit}

\begin{figure}[] 
\centering
\includegraphics [width=0.9\linewidth]{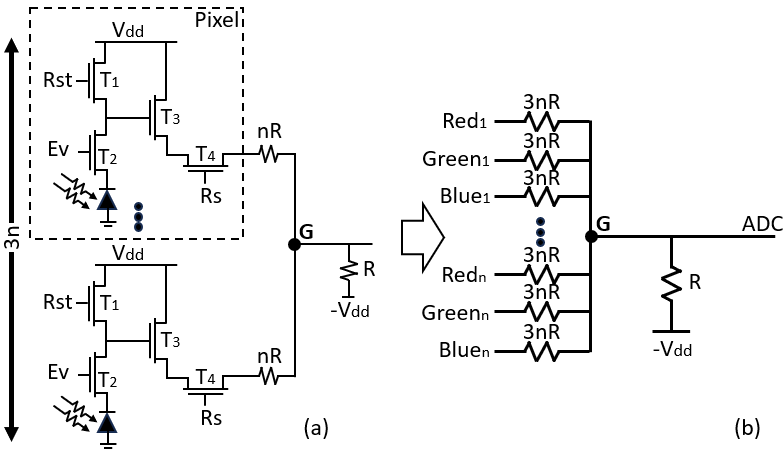}
\vspace{-3mm}
\caption{Analog in-sensor pooling circuit diagram.}
\label{fig:pooling}
\vspace{-4mm}
\end{figure}

Our approach to image compression encompasses two primary steps: grayscale conversion and pooling. These steps are uniquely implemented in the analog domain and occur simultaneously within our proposed compression circuit. The architecture of an individual pixel is depicted in Fig.~\ref{fig:pooling}(a). Within this structure, $T_3$ serves as a source follower (SF), and $T_4$ operates as the row selector (RS) to connect the pixel to the readout circuit. The main idea of our compression approach involves the simultaneous connection of multiple pixels, as illustrated in Fig.~\ref{fig:pooling}(a). For instance, a pooling size of $2 \times 2$ requires the integration of $2 \times 2 \times 3$ pixels, where $3$ represents the RGB channels. Connecting pixels increases the drain voltage of both SF and RS transistors. According to Eq.~\ref{on_condition}, activating a transistor requires the drain-source voltage ($V_{DS}$) to be less than the difference between the gate-source voltage ($V_{GS}$) and the threshold voltage ($V_{TH}$). To solve the issue, we disregard the resistance of SF and RS transistors and assume maximum pixel output ($V_{DD}$). Under this assumption, all pixel resistors (R) effectively operate in parallel. To ensure the voltage of G (in Fig.~\ref{fig:pooling}(a)) remains below zero, we connect them to $-V_{DD}$ through a resistor smaller than R by a factor of $x$, where $x = \frac{R}{\# \text{pixels}}$. For example, when connecting 12 pixels, each pixel's resistor is set to $12R$, and they are collectively connected to $-V_{DD}$ via an $R$ resistor. The simplified equivalent circuit representation is provided in Fig.~\ref{fig:pooling}(b).

\begin{equation}
\small
V_{DS} < (V_{GS} - V_{TH})
\label{on_condition}
\end{equation}

\begin{figure}[] 
\centering
\subfloat[Transient vector for two analog signals]{%
  \includegraphics[clip,width=0.9\columnwidth]{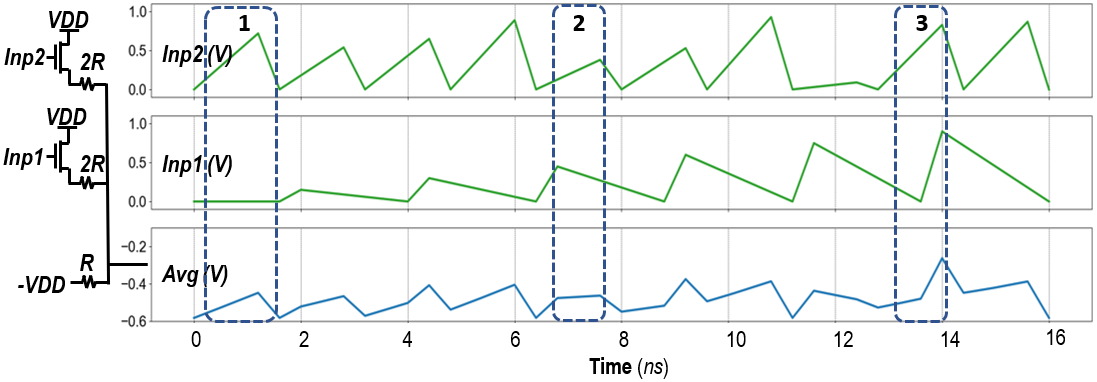}%
}

\subfloat[Transient vector for getting average of 4 digital inputs]{%
  \includegraphics[clip,width=0.9\columnwidth]{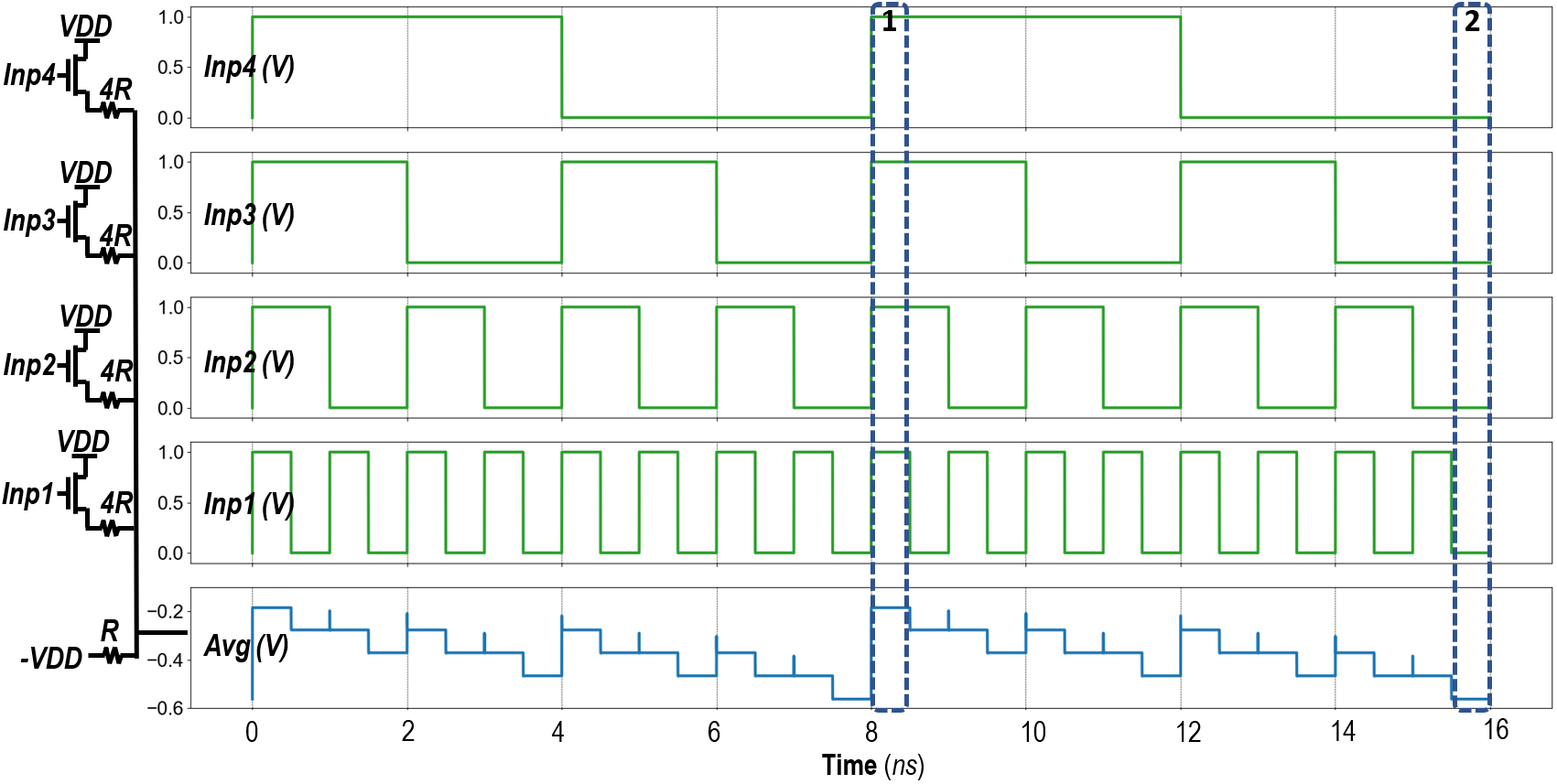}%
}
\vspace{-3mm}
\caption{SPICE simulation of analog compression circuits.}
\label{fig:tran}
\vspace{-6mm}
\end{figure}

To validate the functionality of our proposed circuit, we devised two simplified test benches, as depicted in Fig.~\ref{fig:tran}. The first test bench applies our method to two analog inputs. As illustrated in Fig.~\ref{fig:tran}(a), two transistors are connected via resistors, with their intersection point connected to $-V_{DD}$. The gates of these transistors are connected to two distinct analog voltages, denoted as $Inp_1$ and $Inp_2$. The $Avg$ signal confirms the correct functioning of the circuit. For instance, in \encircle{1}, the $Avg$ signal follows the variations of $Inp_2$ with a more gradual slope, owing to the other input being constant. Another scenario is presented in \encircle{2}, where the input signals have opposing slopes, resulting in an approximately zero slope for the $Avg$ signal. Finally, in \encircle{3}, the influence of $Inp_1$ on the $Avg$ signal can be clearly observed. The second test is depicted in Fig.~\ref{fig:tran}(b). In this configuration, the number of transistors is increased to four, and their gate inputs are connected to digital voltages for enhanced clarity. The $Avg$ signal is observed to follow the average of the inputs precisely. For instance, in \encircle{1}, the $Avg$ signal attains its peak since all inputs are at $V_{DD}$. Conversely, in \encircle{2}, the $Avg$ signal reaches its lowest value, corresponding to the scenario where all inputs are equal to zero. This test bench was extended to accommodate 192 inputs and demonstrated flawless performance.

\begin{table*}[t]
\centering
\caption{Comparison of in-processor (In-Proc) vs in-sensor (In-Sen) scaling. For in-sensor, we use a 2560$\times$1920 pixel array with 8$\times$8, 4$\times$4, and 2$\times$2 pooling yielding the resolutions shown below. For all results, we report the mean average precision (mAP).}
\vspace{-3mm}
\small
\begin{tabular}{lcccccccccccc}
\hline
\multicolumn{1}{c}{Resolution} & \multicolumn{4}{c}{320$\times$240} & \multicolumn{4}{c}{640$\times$480} & \multicolumn{4}{c}{1280$\times$960} \\ \hline
\multicolumn{1}{c}{Color Mode} & \multicolumn{2}{c}{RGB} & \multicolumn{2}{c}{Gray} & \multicolumn{2}{c}{RGB} & \multicolumn{2}{c}{Gray} & \multicolumn{2}{c}{RGB} & \multicolumn{2}{c}{Gray} \\ \hline
\multicolumn{1}{c}{Scaling Mode} & In-Proc & In-Sen & In-Proc & In-Sen & In-Proc & In-Sen & In-Proc & In-Sen & In-Proc & In-Sen & In-Proc & In-Sen \\ \hline
Crowdhuman & 55.2\% & 55.2\% & 52.0\% & 52.0\% & 71.0\% & 71.0\% & 68.1\% & 68.2\% & 79.2\% & 79.5\% & 76.7\% & 76.8\% \\
DHDCampus & 49.6\% & 49.6\% & 49.2\% & 49.2\% & 67.7\% & 67.7\% & 66.8\% & 66.6\% & 80.9\% & 80.9\% & 79.8\% & 79.8\% \\
VisDrone & 19.4\% & 19.3\% & 18.5\% & 18.5\% & 36.8\% & 36.8\% & 34.9\% & 34.7\% & 50.9\% & 50.5\% & 49.7\% & 49.3\% \\ \hline
\end{tabular}
\label{table:accuracy_results}
\vspace{-4mm}
\end{table*}

\section{Results and Discussions}

\vspace{-1mm}
\subsection{Accuracy}
\vspace{-1mm}
To verify that our in-sensor pooling technique does not harm the performance of the stage-1 model, we simulate the in-sensor pooling on three object detection datasets and compare the results to images that are scaled digitally. We also investigate the accuracy/resolution tradeoff, by analyzing three different pooling levels (8$\times$8, 4$\times$4, 2$\times$2) yielding three different image resolutions. Finally, we investigate the effect of our optional grayscale circuit, which can further reduce the memory footprint of images by a factor of 3.

For our evaluation, we use three datasets: Crowdhuman [13], DHDCampus [11], and VisDrone [18]. The Crowdhuman dataset contains pictures of people in large groups (mostly high-resolution) along with bounding boxes for their bodies and heads. DHDCampus contains high-resolution images of people and provides bounding boxes for two classes: person, and cyclist. VisDrone contains high-resolution drone images from above in an urban environment and provides bounding boxes for 10 classes. For all models, we use the YOLOv8 Nano [4] model pre-trained on the MSCOCO [9] object detection dataset. We train our models for 200 epochs and report the final validation mean average precision (mAP) in Table \ref{table:accuracy_results}.

From the table, it can be observed that the in-sensor scaling does not cause a significant accuracy drop. In the majority of cases, we see no drop in accuracy when comparing in-sensor scaling to in-processor scaling. There are a few cases where accuracy is worse and a few cases where accuracy is better. In all cases, higher resolution leads to better accuracy. This is especially true in the case of VisDrone where the accuracy more than doubles between the smallest and highest resolution. This dataset is likely the most sensitive to image size because the images are taken from far away, thus making it difficult to find objects in lower resolution. In our experiments for grayscale, the images that were trained in RGB and evaluated in grayscale (not shown in the table) suffered a noticeable accuracy drop of 3.4\%-6.7\%. To mitigate this drop in accuracy, we retrain the models using grayscale images (shown in the table) which reduces the accuracy drop to around 0.4\%-3.2\%. For grayscale, Crowdhuman is the most sensitive, while the other datasets are only slightly affected.

\vspace{-3mm}
\subsection{Memory Utilization}
\begin{figure}[t]
\begin{minipage}{\columnwidth}
\centering
\subfloat[In-Processor Scaling]{%
  \includegraphics[clip,width=0.5275\linewidth, height=1.5in]{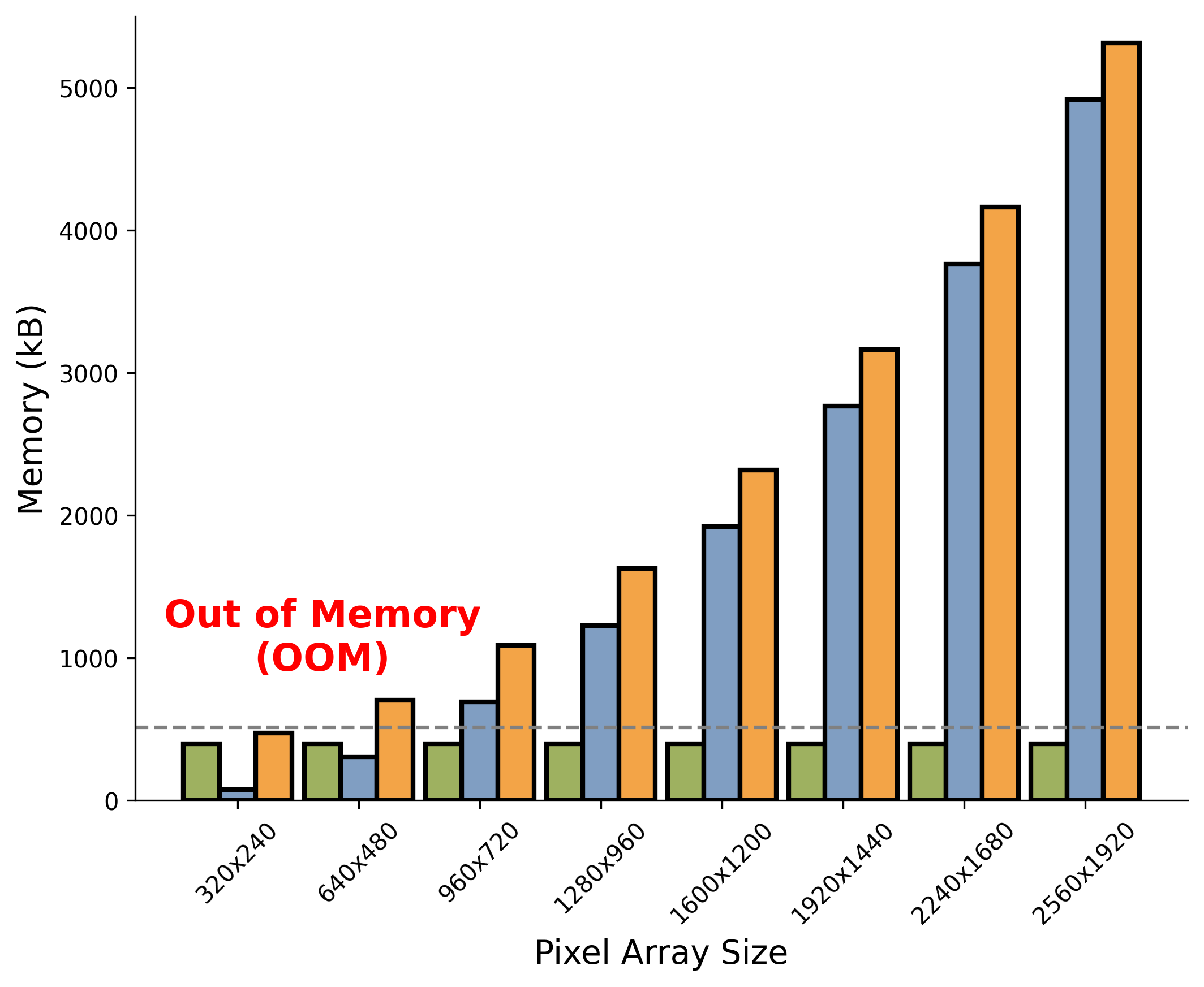}%
}\hfill
\subfloat[In-Sensor Scaling]{%
  \includegraphics[clip,width=0.4725\linewidth, height=1.5in]{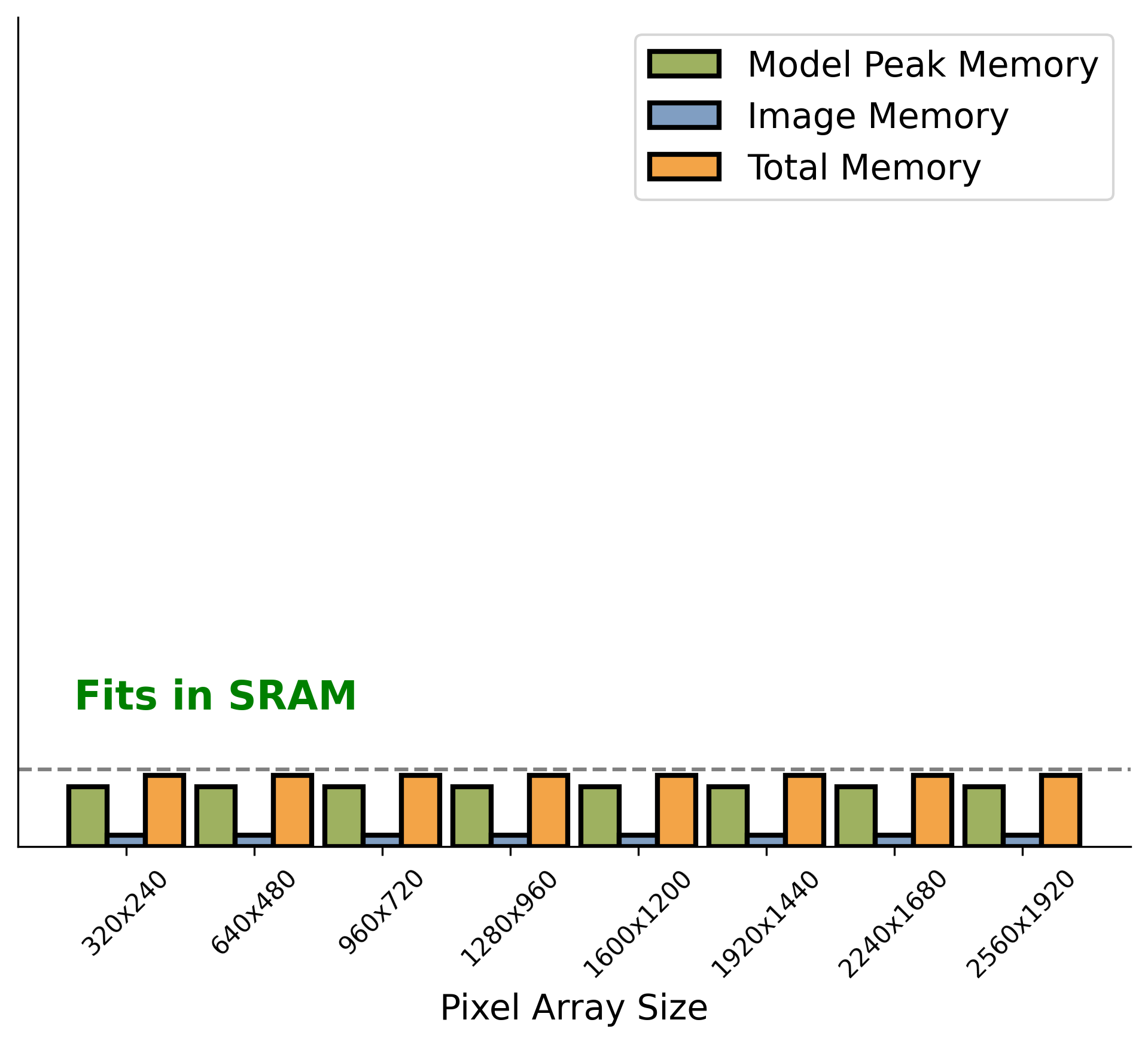}%
}
\vspace{-3mm}
\caption{Comparison of memory requirements for the two-stage model with varying resolutions along with peak memory of stage-1 and stage-2 models. We scale images down to 320$\times$240 pixels for the stage-1 model in both cases.}
\label{fig:memory_requirements}
\end{minipage}
\vspace{-6mm}
\end{figure}
As discussed earlier, many edge devices have strict limits on the amount of memory available. 
While specific memory requirements depend on the exact hardware and application, as a case study we examine an extreme scenario of running a two-stage system with an object detection model and recognition model on a microcontroller.

We use the popular STM32H743 as our hardware of choice, which uses the Arm Cortex M7 and has 512kB SRAM/2MB flash memory. To run a two-stage system on this hardware we need the peak activation memory of each model to be below 512kB and the total weights of both models to be below 2MB. We pick two off-the-shelf tiny ML models and analyze peak SRAM and flash memory utilization. We use the MCUNetV2 person detection model [7] as our stage 1 object detection model along with the MCUNetV2 image classification model [7] as our stage 2 model.

We calculate the total memory usage for both models in terms of peak SRAM and flash memory. We use TFLite-Micro as our interpreter and analyze peak SRAM by looking at the execution order of operations for both models and finding the point where the most memory is required. To calculate the flash memory utilization, we add up the weight memory for each model. For the stage 1 model, we find 337kB/296kB peak SRAM/flash usage and for the stage 2 model, we find 398kB/1MB peak SRAM/flash usage for a total of 398kB peak SRAM and 1.3MB flash. This leaves only 114kB of SRAM free during the inference of either model.

If we use in-processor scaling, 114kB can be enough for the stage-1 model to find ROI. However, as discussed earlier and shown in Fig. \ref{fig:detection_example}, the resulting ROIs will be too small for the stage-2 model to extract the rich features it relies on. As shown in Figure \ref{fig:memory_requirements} (a), the system quickly runs out of memory using in-processor scaling if the pixel array size increases. On the other hand, the HiRISE can support large pixel arrays since the high-resolution pixels are kept off the digital hardware. Using scaling, the stage-1 image can be kept under 114kB. Finally, when the high-resolution ROI is needed for the stage-2 model, only the pixels related to that ROI are extracted. Since the full-resolution image never leaves the analog domain, the size of the image no longer exceeds our memory requirements using pooling and selective ROI, as shown in Fig. \ref{fig:memory_requirements} (b).

\subsection{Data Transfer}
\begin{figure}[] 
\centering
\includegraphics [width=0.98\linewidth]{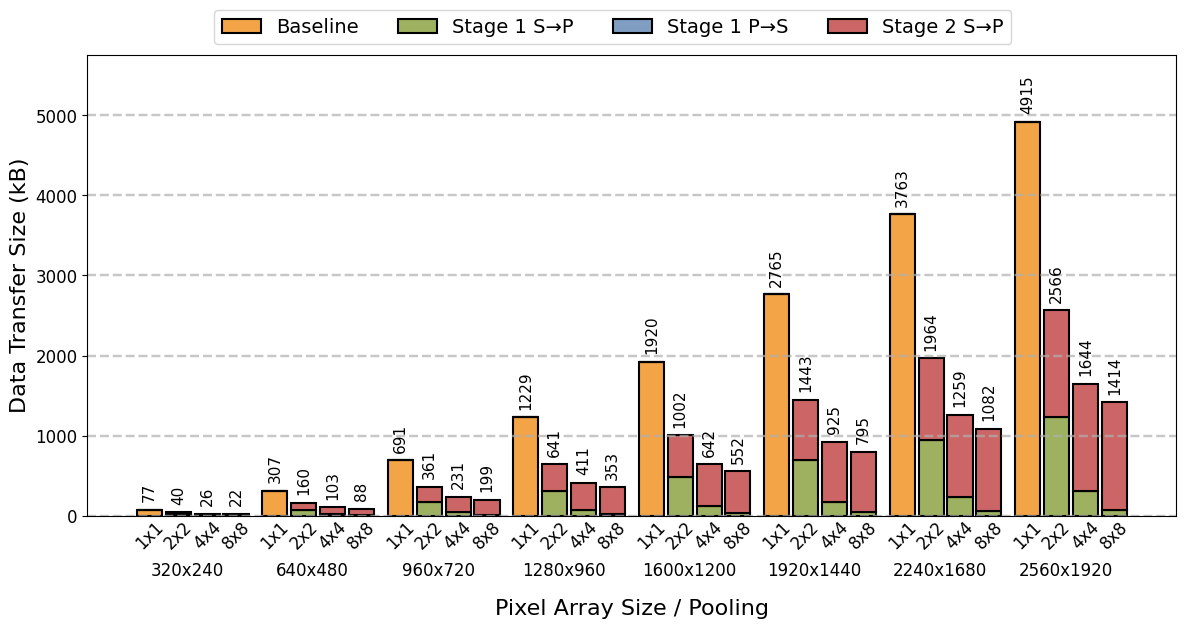}
\vspace{-5mm}
\caption{Data transfer requirements for different pixel array sizes with different pooling levels.}
\vspace{-6mm}
\label{fig:data_transfer}
\end{figure}

Although the peak memory using in-sensor scaling and in-sensor ROI is much lower, the data must go back and forth between the sensor and the processor multiple times. To investigate the data transfer requirements of our system, we evaluate the median data transfer. In addition, we evaluate the median packet size, i.e., width and height of ROI, at various resolutions for each dataset. We find that the Crowdhuman dataset has the largest total data transfer size. We report our findings in Figure \ref{fig:data_transfer}. As shown in the figure, even though there is more back and forth between the sensor and processor using the HiRISE system, the total data transfer size is smaller in all cases. 

For 2$\times$2 pooling, there is a 1.9$\times$ reduction in data transfer over the baseline. Sending the initial pooled image from the camera to the digital hardware ($D1_{S\rightarrow P}$) accounts for 48\% of the data transfer, while sending the cropped images for each bounding box from the camera to the digital hardware ($D2_{S\rightarrow P}$) accounts for 52\% of the data transfer. As mentioned in Section 2, the data transfer required for sending the dimensions and coordinates of ROI from processor to sensor ($D1_{P\rightarrow S}$) is negligible compared to $D1_{S\rightarrow P}$ and $D2_{S\rightarrow P}$. For 4$\times$4 pooling we have a 3$\times$ reduction in data transfer over the baseline. $D1_{S\rightarrow P}$ accounts for 19\% of the data transfer while $D2_{S\rightarrow P}$ accounts for 81\% of the data transfer. Finally, for 8$\times$8 pooling, we have a 3.5$\times$ reduction in data transfer over the baseline. $D1_{S\rightarrow P}$ accounts for just 5\% of the data transfer, while $D2_{S\rightarrow P}$ accounts for 95\% of the data transfer. These ratios hold for all resolutions, and the total data transfer for different resolutions broken down by each stage can be seen in Fig. \ref{fig:data_transfer}. Thus, although HiRISE requires moving the data from the sensor to the processor twice, using in-sensor pooling along with in-sensor ROI selection, it can achieve a significant reduction in the overall data transfer in all cases.

\vspace{-3mm}
\subsection{Energy Consumption}
\vspace{-1mm}
Although all the reduction in memory and data transfer corresponds directly to savings in energy
, these are not the only sources of energy saving in the HiRISE system. Since the HiRISE system enables converting fewer pixels from analog to digital, it requires fewer ADCs which are a major source of energy consumption in sensors. To evaluate the energy savings of HiRISE, we use HSPICE and Synopsys Design Compiler to implement the analog compression/conversion units and digital control circuitry, respectively, using 45nm transistor technology and 45nm 8-bit ADC [3]. 

\begin{figure}[] 
\centering
\includegraphics [width=0.99\linewidth]{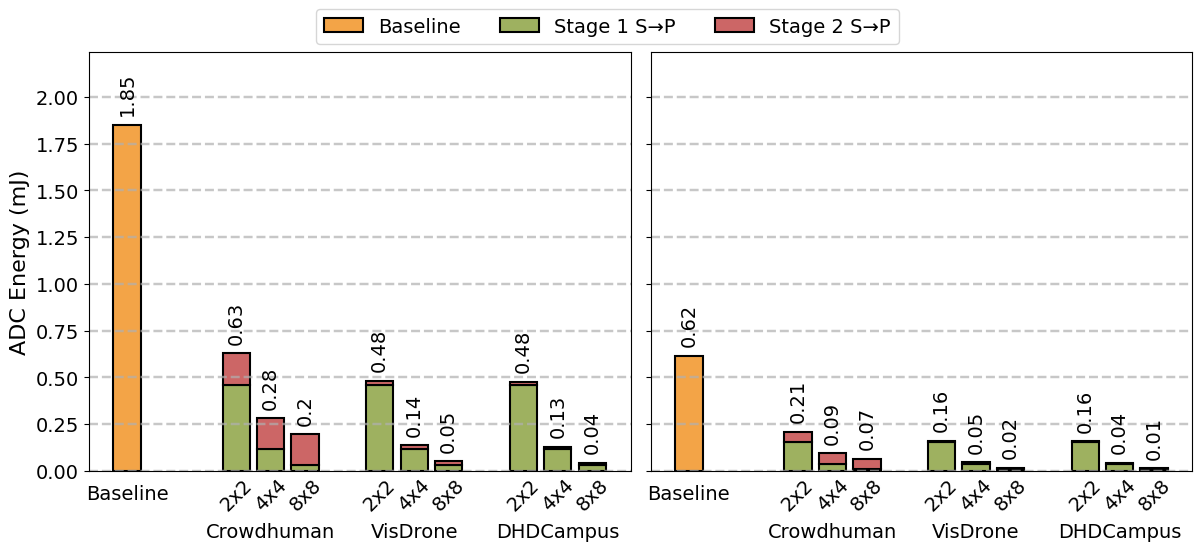}
\vspace{-3mm}
\caption{Energy consumption under different pooling levels with RGB (left) and grayscale (right).}
\label{fig:energy}
\vspace{-6mm}
\end{figure}

Figure \ref{fig:energy} shows the energy consumption of HiRISE for different pooling sizes for both RGB and grayscale images across three datasets. For our experiments, we use a pixel array size of 2560$\times$1920 for all datasets. In the figure, the baseline system is a 1-stage system that converts the entire image and transfers it to the processor for the end-goal task. The ADC energy consumption of stage 2 in the HiRISE system varies for different images depending on the size and number of ROIs detected in stage 1. Therefore, for each bar shown in Fig. \ref{fig:energy}, we measure the median energy consumption for the entire dataset except for the baseline which has a fixed energy consumption of 1.85mJ per image. 
As shown, the Crowdhuman dataset exhibits higher overall energy consumption compared to the other two datasets. This is attributed to the substantial quantity of objects present in each image, resulting in an increased intersection over the union of ROI boxes. 

For the Crowdhuman dataset, a HiRISE system with 2$\times$2 pooling can achieve roughly 3$\times$ energy reduction compared to the baseline with 0.63mJ energy consumption. Of the 0.63mJ, 73\% of the energy (0.46mJ) comes from converting the initial pooled image (Stage 1), while 27\% of the energy is from converting pixels from the ROI (Stage 2). For 4$\times$4 pooling using the same dataset, a 6.5$\times$ reduction is observed with 0.28mJ total energy consumption. Of the 0.28mJ, 41\% of the energy (0.12mJ) consumption comes from stage 1, while 59\% of the energy is consumed for converting ROI pixels. Finally, for 8$\times$8 pooling, a 9.4$\times$ reduction is realized over the baseline model with 0.2mJ energy consumption. Of the 0.2mJ, 15\% of the energy (0.03mJ) is consumed for converting the pooled image, while 85\% is consumed in stage 2. Across all experiments, the energy consumption of analog pooling circuitry varies between 1.71 nJ and 91.4 nJ which is several orders of magnitude smaller than ADC conversion, thus having a negligible impact on the overall energy.  

\begin{table*}[]
\small
\centering
\caption{End-to-end system analysis for different pixel array sizes using 320$\times$240 pooling resolution.}
\vspace{-4mm}
\begin{tabular}{ccccccccccccc}
\hline
\multirow{3}{*}{Model} & \multirow{3}{*}{Pixel Array} & \multirow{3}{*}{ROI} & \multirow{3}{*}{Acc\%} & \multicolumn{5}{c}{SRAM (kB)} & \multicolumn{2}{c}{\multirow{2}{*}{Data Transfer (kB)}} & \multicolumn{2}{c}{\multirow{2}{*}{Energy (mJ)}} \\ \cline{5-9}
 &  &  &  & \multirow{2}{*}{Peak Act} & \multicolumn{2}{c}{Image Memory} & \multicolumn{2}{c}{Total} & \multicolumn{2}{c}{} & \multicolumn{2}{c}{} \\ \cline{6-9}
 &  &  &  &  & Baseline & HiRISE & Baseline & HiRISE & Baseline & HiRISE & Baseline & HiRISE \\ \hline
\multirow{8}{*}{\rotatebox[origin=c]{90}{MCUNetV2}} & 320$\times$240 & 14$\times$14 & 58.6 & 6.4 & 230 & \multirow{8}{*}{230} & 237 & 237 & 230 & 240 & 0.029 & 0.030 \\
 & 640$\times$480 & 28$\times$28 & 71.5 & 14.3 & 922 &  & 936 & 245 & 922 & 268 & 0.115 & 0.034 \\
 & 960$\times$720 & 42$\times$42 & 76.8 & 28.1 & 2,074 &  & 2,102 & 258 & 2,074 & 315 & 0.259 & 0.039 \\
 & 1280$\times$960 & 56$\times$56 & 78.0 & 46.6 & 3,686 &  & 3,733 & 277 & 3,686 & 381 & 0.461 & 0.048 \\
 & 1600$\times$1200 & 70$\times$70 & 80.8 & 69.7 & 5,760 &  & 5,830 & 300 & 5,760 & 466 & 0.720 & 0.058 \\
 & 1920$\times$1440 & 84$\times$84 & 80.3 & 97.6 & 8,294 &  & 8,392 & 328 & 8,294 & 569 & 1.037 & 0.071 \\
 & 2240$\times$1680 & 98$\times$98 & 81.1 & 130 & 11,290 &  & 11,420 & 361 & 11,290 & 691 & 1.411 & 0.086 \\
 & 2560$\times$1920 & 112$\times$112 & 81.2 & 168 & 14,746 &  & 14,913 & 398 & 14,746 & 833 & 1.843 & 0.104 \\ \hline
\multirow{8}{*}{\rotatebox[origin=c]{90}{MobileNetV2}} & 320$\times$240 & 14$\times$14 & 65.5 & 12.5 & 230 & \multirow{8}{*}{230} & 242 & 243 & 230 & 240 & 0.029 & 0.030 \\
 & 640$\times$480 & 28$\times$28 & 72.9 & 43.4 & 921 &  & 964 & 274 & 922 & 268 & 0.115 & 0.034 \\
 & 960$\times$720 & 42$\times$42 & 78.4 & 93.1 & 2,074 &  & 2,167 & 324 & 2,074 & 315 & 0.259 & 0.039 \\
 & 1280$\times$960 & 56$\times$56 & 81.0 & 161 & 3,686 &  & 3,848 & 392 & 3,686 & 381 & 0.461 & 0.048 \\
 & 1600$\times$1200 & 70$\times$70 & 82.4 & 249 & 5,760 &  & 6,009 & 479 & 5,760 & 466 & 0.720 & 0.058 \\
 & 1920$\times$1440 & 84$\times$84 & 83.6 & 355 & 8,294 &  & 8,650 & 586 & 8,294 & 569 & 1.037 & 0.071 \\
 & 2240$\times$1680 & 98$\times$98 & 84.4 & 480 & 11,290 &  & 11,770 & 711 & 11,290 & 691 & 1.411 & 0.086 \\
 & 2560$\times$1920 & 112$\times$112 & 84.7 & 624 & 14,746 &  & 15,367 & 854 & 14,746 & 833 & 1.843 & 0.104 \\ \hline
\end{tabular}
\label{tab:end_to_end_stage2}
\vspace{-4mm}
\end{table*}

\vspace{-3mm}
\subsection{End-to-End System}
\vspace{-1mm}
Here, we provide a comprehensive analysis of the end-to-end system using the  Real-world Affective Faces Database (RAF-DB) dataset. We compare results across two models, i.e., MobileNetV2 [12] and MCUNetV2 [7], across different pixel array sizes. For each pixel array size, we find the average ROI by analyzing over 100,000 ROIs for head detection from the Crowdhuman dataset. Using these ROIs, we train a facial expression recognition model on the RAF-DB dataset. We train for 200 epochs using the same approach as POSTERV2 [10]. For each of the models, we report accuracy, peak memory of the entire system, and data transfer and show the effect of using HiRISE. We also show the energy consumption of the sensor both with and without HiRISE. In all cases for HiRISE, we use pooling such that the output resolution for the stage-1 model is 320$\times$240. 

From Table \ref{tab:end_to_end_stage2} it can be observed that for a pixel array size of 320$\times$240, the detected ROI size becomes very small (14$\times$14). Since there is not much information remaining for the stage-2 model, the models struggle to recognize facial expressions, with accuracies of 58.6\% and 65.5\% for MCUNetV2 and MobileNetV2, respectively. The peak SRAM, data transfer, and energy at this pixel array size are similar for both systems. For the pixel array size of 640$\times$480, the accuracy improves significantly for both models with 71.5\% and 72.9\% accuracy for MCUNetV2 and MobileNetV2, respectively. This indicates that they can extract more high-level features with a larger ROI. Simultaneously, HiRISE utilizes 3.8$\times$ less SRAM than the baseline and 3.4$\times$ less energy in-sensor. When the pixel array size is doubled to 1280$\times$960, an accuracy of 78\% and 81\% could be realized for MCUNetV2 and MobileNetV2, respectively. For this array size using MCUNetV2, we see that HiRISE achieves a 13.5$\times$ and 9.7$\times$ reduction in SRAM utilization and energy consumption compared to the baseline, respectively. Finally, for the maximum pixel array size investigated, i.e., 2560$\times$1920, an 81\% and 84.7\% accuracy is obtained for MCUNetV2 and MobileNetV2, respectively. At this array size, HiRISE uses 37.5$\times$ less SRAM than the baseline for MCUNetV2, while achieving a 17.7$\times$ reduction in energy consumption.

\vspace{-4mm}
\section{Conclusion}
\vspace{-1mm}
In this paper, we proposed HiRISE, a system that enables the deployment of two-stage models that process high-resolution images at resource- and energy-constrained edge devices. HiRISE achieves this by ensuring that only the necessary information leaves the sensor. This is made possible by designing a novel in-sensor pooling circuit along with a selective ROI mechanism. Through comprehensive end-to-end experiments, we have demonstrated that HiRISE not only significantly reduces the memory demands of edge ML applications, but also realizes considerable energy savings and reduction in the total data transfer compared to conventional systems.

\vspace{-12pt}
\section*{Acknowledgment}
\vspace{-3pt}
This work is supported in part by the National Science Foundation (NSF) under grant numbers 2340249, 2216773, 2228028, and 2216772. 
\vspace{-20pt}
\bibliographystyle{ACM-Reference-Format}
\bibliography{Reference.bib}

\end{document}